\documentclass[pdflatex,sn-mathphys-num]{sn-jnl}

\usepackage{multirow}%
\usepackage{amsmath,amssymb,amsfonts}%
\usepackage{amsthm}%
\usepackage{mathrsfs}%
\usepackage[title]{appendix}%
\usepackage{xcolor}%
\usepackage{textcomp}%
\usepackage{manyfoot}%
\usepackage{booktabs}%
\usepackage{algorithm}%
\usepackage{algorithmicx}%
\usepackage{algpseudocode}%
\usepackage{listings}%
\usepackage{graphicx}
\usepackage{tabularx}
\usepackage{multirow}
\usepackage{subcaption}
\usepackage{bm}

\usepackage{xspace}
\makeatletter
\DeclareRobustCommand\onedot{\futurelet\@let@token\@onedot}
\def\@onedot{\ifx\@let@token.\else.\null\fi\xspace}
\def\eg{\emph{e.g}\onedot} 
\def\ie{\emph{i.e}\onedot} \def\Ie{\emph{I.e}\onedot}
\def\cf{\emph{c.f}\onedot} 
\def\etc{\emph{etc}\onedot} 
\def\wrt{w.r.t\onedot}

\DeclareMathOperator*{\argmin}{argmin}

\newcommand{\beginsupplement}{%
        \setcounter{table}{0}
        \renewcommand{\thetable}{S\arabic{table}}%
        \setcounter{figure}{0}
        \renewcommand{\thefigure}{S\arabic{figure}}%
     }
     
\theoremstyle{thmstyleone}%
\theoremstyle{thmstyletwo}%

\theoremstyle{thmstylethree}%

\begin{document}

\title[Automating Catheterization Labs with Real-Time Perception]{Automating Catheterization Labs with Real-Time Visual Perception}


\author[1]{\fnm{Fan} \sur{Yang}}\email{fan.yang03@uii-ai.com}
\equalcont{These authors contributed equally to this work.}
\author[2]{\fnm{Benjamin} \sur{Planche}}\email{benjamin.planche@uii-ai.com}
\equalcont{These authors contributed equally to this work.}
\author[2]{\fnm{Meng} \sur{Zheng}}\email{meng.zheng@uii-ai.com}
\author[2]{\fnm{Cheng} \sur{Chen}}\email{cheng.chen05@uii-ai.com}
\author[2]{\fnm{Terrence} \sur{Chen}}\email{terrence.chen@uii-ai.com}
\author*[2]{\fnm{Ziyan} \sur{Wu}}\email{ziyan.wu@uii-ai.com}

\affil[1]{\orgname{United Imaging Intelligence},  \city{Shanghai},  \country{China},\postcode{200232}}

\affil*[2]{\orgname{United Imaging Intelligence},  \city{Burlington}, \state{MA}, \country{USA},\postcode{01803} }


\abstract{For decades, three-dimensional C-arm Cone-Beam Computed Tomography (CBCT) imaging system has been a critical component for complex vascular and nonvascular interventional procedures. While it can significantly improve multiplanar soft tissue imaging and provide pre-treatment target lesion roadmapping and guidance,  the traditional workflow can be cumbersome and time-consuming, especially for less experienced users. To streamline this process and enhance procedural efficiency overall, we proposed a visual perception system, namely \textit{AutoCBCT}, seamlessly integrated with an angiography suite. This system dynamically models both the patient's body and the surgical environment in real-time. \textit{AutoCBCT} enables a novel workflow with automated positioning, navigation and simulated test-runs, eliminating the need for manual operations and interactions. The proposed system has been successfully deployed and studied in both lab and clinical settings, demonstrating significantly improved workflow efficiency.}




\maketitle

\section{Introduction}
\label{sec:introduction}
\begin{figure*}[t]
	\centering
	\includegraphics[width=1\textwidth]{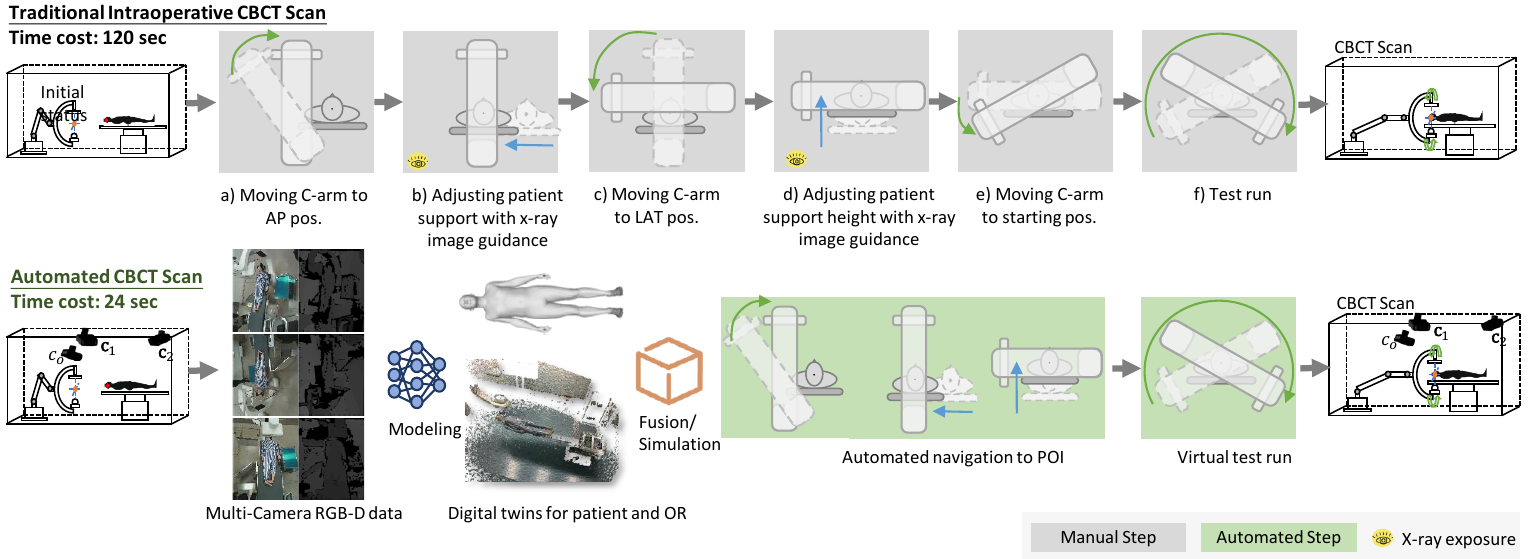}
	\caption{Left: Conventional C-arm CBCT of the head 
	(red dot: region of interest; orange dot: radiation center; green arrows: C-arm motion).
	Right: (A) Vision-empowered C-arm system in an operating room (with $\mathbf{c}_0$, $\mathbf{c}_1$, $\mathbf{c}_2$  RGB-depth sensors). (B) Paired RGB-D images obtained from the cameras. (C) Resulting fused point cloud displayed from three views. 
	}
	\label{fig:c-arm_overview}
\end{figure*}

C-arm Cone-Beam Computed Tomography (CBCT) utilizes an advanced imaging setup comprising a flat panel detector and an X-ray tube unit, each positioned on opposite ends of a C-shaped robotic arm (Fig.~\ref{fig:c-arm_overview}, left). This configuration allows the system to offer a range of functions including projection radiography, fluoroscopy, digital subtraction angiography, and volumetric computed tomography (CT); all within a single patient setup~\cite{orth2008c} during interventions. Thanks to this versatile capability, interventional radiologists can conduct intraprocedural volumetric imaging in one operating room without the need for patient transportation~\cite{orth2008c}. Moreover, C-arm CBCT significantly enhances the safety and efficacy of both complex vascular and non-vascular interventional procedures~\cite{wallace2008three}.
While the emergence of C-arm CBCT greatly enhances interventional procedures and instills a subjective level of confidence among clinicians, the CBCT workflow remains complex for inexperienced operators. Typically, this workflow involves two steps: patient positioning and test-runs. During patient positioning, the user manually adjusts the C-arm to focus the X-ray tube onto the target organ/tissue. Using guidance from the X-ray images, the user refines the positioning by adjusting either the C-arm or the patient support so that the central beam axis of the X-ray tube intersects with the center of the target organ/tissue. 
The primary goal of the following test-run is to verify that the scan can proceed without any collisions. The C-arm rotates according to a specified protocol, and if a collision is detected during the test-run, the user must either re-position the patient support or remove any obstacles. This process often requires multiple iterations and can consume substantial treatment time before the actual intraoperative scan can take place (see Fig. \ref{fig:c-arm_overview},left). Furthermore, each time the patient is re-positioned, additional X-ray imaging is necessary, resulting in increased radiation exposure for the patient, as well as clinicians around.

With advancements in optical sensors and computer vision, recent efforts have aimed at optimizing the CBCT workflow. These include reconstructing the interventional room in 3D \cite{Ladikos2008RealTime3R}, real-time monitoring of the C-arm's surroundings with 3D cameras \cite{Incetan2019}, and automatic localization of major body parts of the patient \cite{Schaller09}. While these methods optimizes individual steps of the system by enhancing, to some extent, its perception capabilities; they do not completely eliminate the need for manual intervention.

To address the aforementioned issues, we introduce, to the best of knowledge, the \textit{first} vision-based system, \textit{AutoCBCT}, designed to facilitate fully-automated C-arm CBCT scanning. This system comprises three main components:
(1) a multi-view visual sensing module equipped with multiple RGB-D cameras;
(2) a real-time 3D patient modeling module; 
and (3) a virtual test run (\textit{VTR}) module.
The visual sensing module captures synchronized RGB and depth images from multiple viewpoints, providing a comprehensive depiction of the surgical environment and patient. 
Using these real-time, multi-view, multi-modal images, the 3D patient modeling function models the patient by recovering the SMPL 3D pose and shape parameters~\cite{loper2015smpl}. This process implies: (a) the detection of salient body keypoints in each RGBD frame; (b) their robust triangulation to regress their position in the global 3D coordinate system; (c) their consolidation over time to handle transient occlusions and other environmental challenges; and (d) the final regression of the patient's body shape and pose information according to the upstream predictions.
Benefiting from a variety of technical contributions detailed in this article, this complex machine-learning-based module enables the 3D localization of specified anatomical regions. 
By integrating the estimated location of the target body region into the C-arm robot system, the system autonomously plans trajectories and efficiently navigates the robot arm to the target location, all without the need for X-ray exposures.

Once the C-arm is positioned, the virtual test run module utilizes images from the visual sensor module to reconstruct the environment and integrate it with the 3D patient model. Using the planned scanning trajectory and engineering data of the robotic system (\eg, CAD files of the C-arm), the \textit{VTR} module simulates the movement of the C-arm within the virtual environment alongside the reconstructed surgical environment and patient model. This simulation allows for efficient detection of potential collisions with the patient, devices, or other obstacles, which can then be visualized and presented to users. Users can then take appropriate actions (\eg, removing the obstacles) to ensure a smooth and efficient scan, eliminating the need for lengthy manual trials.

In summary, our work makes the following contributions:
\begin{enumerate}
    \item Development of the first vision-based C-arm system capable of fully automated C-arm CBCT, significantly enhancing workflow efficiency.
    \item Introduction of a novel multi-view 3D patient body modeling method that estimates the 3D shape and pose of the patient, and infers the location of predefined anatomical regions through fusion with multiple RGBD cameras.
    \item Design of a novel virtual test run module, eliminating the need for manual test runs.
    \item Extensive evaluations conducted in both laboratory and clinical environments, demonstrating the efficacy and efficiency of the proposed system.
\end{enumerate}

\section{Results}

\subsection{Clinical Value}\label{sec:clinical_val}
Adopting a top-down approach to evaluating the proposed system, we first demonstrate the clinical value of the proposed automated CBCT workflow in interventional settings, considering a variety of domain-relevant metrics.

\subsubsection{Experimental Protocol}

To demonstrate the effectiveness, efficiency, and safety of the system and its core components for clinical applications, all experiments are conducted in both lab and real clinical environments. All experiments/data obtained approvals from ethical review boards.

\paragraph{System Deployment}
The proposed vision-based guidance solution is tightly integrated into two angiography C-arm systems, deployed in both laboratory and hospital-based interventional suites. Both of the systems, including a floor-mounted robotic angio system and a ceiling-mounted angio system, went through multi-site clinical trials and are certified by National Medical Products Administration (NMPA) of China. Since the NMPA clearance, these systems have been clinically deployed and are being actively used in more than 5 hospitals. Multi-site clinical evaluation results are discussed in Section \ref{sc:eval}. The robotic module of the floor-mounted angio system has 
The 3D patient body modeling module is running on Nvidia TX2 with 8GB of shared memory. The virtual test run module is running on a console PC with Intel Xeon Silver 4210 processor.

\begin{figure*}[t]
	\centering
	\includegraphics[width=1\textwidth]{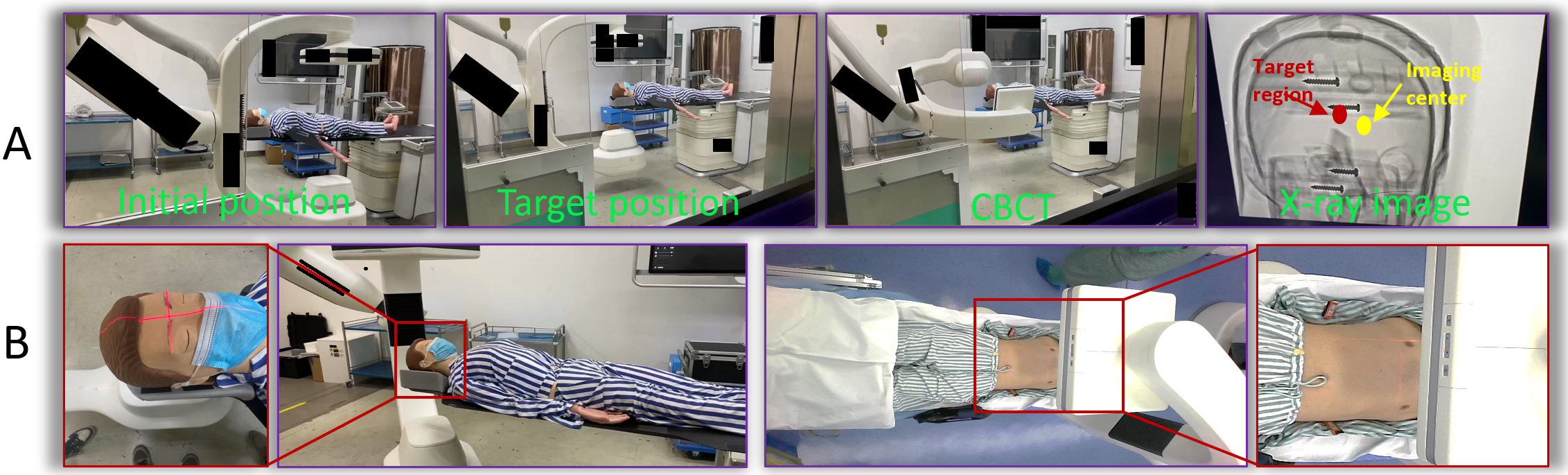}
	\caption{(A) Test protocol of C-arm CBCT. (B) Conventional patient positioning.}
	\label{fig:cbct_protocol}
\end{figure*}

\paragraph{Data Collection}
\noindent
\textbf{End-to-end \textit{AutoCBCT}.}  
The protocol to testing our entire automated CBCT system against conventional CBCT is depicted in Fig.~\ref{fig:cbct_protocol} A. Scanner positioning \wrt the user-predefined anatomical region of interest (ROI) is either controlled through the operator's manual inputs (manual mode) or guided by the estimated 3D patient body model (auto), and is deemed successful if the distance between the final imaging center and target region is under a clinical threshold ($<$25mm). Note that the initial position of the C-arm can be any possible location in the surgical space. 

The lab dataset is collected with diverse environmental settings including 3 different patient support positions, and 3 initial C-arm positions.
The clinical data covers 26 subjects from two different hospitals. Among them, 16 subjects are scanned with \textit{AutoCBCT}, and the other 10 samples were with conventional CBCT as contrast group. Note that samples are acquired \wrt head scan protocol.  

\noindent
\textbf{Patient positioning.} 
The proposed 3D patient modeling components of \textit{AutoCBCT} are tested by integrating them into a patient positioning procedure, where  the C-arm is first moved to its initial position then navigated to the target region manually or by the 3D patient modeling module. 

The patient positioning procedure is leveraged to test the 3D patient modeling module in a system perspective. Following conventional approach (\cf Fig.~\ref{fig:cbct_protocol} B), the C-arm flat panel is moved to the target region by using laser light to align with the ROI. During this process, X-ray imaging is jointly utilized to assist the fine ROI localization. In contrast, the 3D patient modeling yields the estimated 3D location of the ROI, enabling the automated navigation of the C-arm robot to the target region.  

Lab data is acquired by repeating this procedure 10 times. For clinical evaluation, samples of 34 subjects are obtained by moving the C-arm from chest to right radial artery (10 samples) and from head to right radial artery (24 samples), accounting for diverse patient occlusions scenarios and support positions. 

\noindent
\textbf{Virtual test run (\textit{VTR}).} 
As for the simulated test run component, we evaluate its applicability to clinical scenarios by randomly placing various common equipment/devices around the C-arm. \textit{VTR} and manual test run are performed respectively to clear the scanning path. 
Fig. \ref{fig:c-arm_overview} illustrates the steps of a conventional C-arm CBCT test run. The conventional workflow consist of 6 steps, two of which require x-ray exposures to guide the adjustment of patient support. With the proposed multi-camera perception system and the fully automated workflow, the entire test run process can be done with one click, and typically no x-ray exposure is required. 
\textit{VTR} is evaluated in lab settings over 10 different scenarios (including head-side-vertical, full-head-vertical, left-side-vertical, right-side-vertical full-left vertical, full-right vertical etc.) containing various objects, all with head-scanning protocol; as well as jointly with the auto-positioning module on the corresponding clinical data.  


\paragraph{Metrics}

Comparison metrics include in-position time (ArT), \ie, time from scan initiation to C-arm final positioning; interaction times (ITs), \ie, number of times that the user needs to interactively control the C-arm during patient positioning and CBCT.
We also measure the one-pass successful rate (OSR), \ie, the percentage of successfully accomplished CBCT or patient positioning without any interference, from initial position to completing CBCT or to target position. The success is measured in terms of absolute distance between target region and image region ($<$25mm). 
X-ray imaging time (XIT) and X-ray imaging dose (XID) respectively correspond to the imaging time and radiation dose that are involved before actual CBCT. 
More specific to \textit{VTR}, we evaluate the collision detection precision (CDP), \ie, a precision score calculated by comparing the detected collision regions by \textit{VTR} with the recorded ones from actual test run; and the test run time (TRT), \ie, time to complete virtual test run or actual test run.

\subsubsection{Evaluation in Lab and Clinical Settings}\label{sc:eval}

\paragraph{\textit{AutoCBCT}}
Table~\ref{table:AutoCBCT} shows that \textit{AutoCBCT} outperforms conventional C-arm CBCT (C$^3$BCT) on both lab and clinical tests. The lower ArT and XIT are evidence of the substantially higher efficiency of \textit{AutoCBCT}, whereas its 2/3 decrease in ITs demonstrates superior usability. As indicated by the zero value assigned to XID, no additional imaging radiation is released before actual CBCT scan when employing our system.
These results highlight that, compared to C$^3$BCT, \textit{AutoCBCT} significantly reduces the preparation time of image acquisition (from 1.5min to 25s) and simplifies the operators' workflow, while maintaining high precision (100$\%$ OSR). Qualitative results are shared in Fig.~\ref{fig:vis} left.



\begin{table}[!t]
\small
\center
\scalebox{0.95}{
\begin{tabular}{c|c|c|c|c|c}
\toprule
                  
                  \textbf{Lab} & ArT (s)$\downarrow$ &ITs$\downarrow$ & OSR$\uparrow$&XIT (s)$\downarrow$&XID (mGy)$\downarrow$\\ \hline\hline
                  C$^3$BCT&  92.58 &9 &0&11.36& 2  \\  
                  \textbf{AutoCBCT}&\textbf{24.61} & \textbf{3}& \textbf{100\%}&\textbf{0}&\textbf{0}  \\ \bottomrule
\end{tabular}}
~
\scalebox{0.95}{
\begin{tabular}{c|c|c|c|c|c}
\toprule
                  
                  \textbf{Clinical} & ArT (s)$\downarrow$ &ITs$\downarrow$ & OSR$\uparrow$&XIT (s)$\downarrow$&XID (mGy)$\downarrow$\\
                  \midrule
                  C$^3$BCT&  88.90&9 &0&10.20& 2.4  \\  
                  \textbf{AutoCBCT}&  \textbf{23.82}&\textbf{3} &\textbf{100\%} &\textbf{0} &\textbf{0}  
                  \\ \bottomrule
\end{tabular}}
\caption{Clinical evaluation of \textit{AutoCBCT} and conventional C-arm CBCT (C$^3$BCT).}  
\label{table:AutoCBCT}
\end{table}

\paragraph{Patient Positioning Module}
As shown in Table~\ref{table:3Dkeypoints}, in lab settings, the computer-vision-based auto positioning and C-arm navigation (\ie, auto-positioning) surpasses conventional manual operations in all metrics over the two selected regions of interest (right femoral artery and right radial artery). The auto-positioning requires 2 ITs, while conventional operations require about 7 to 9 ITs. This demonstrates that auto-positioning can significantly simplify this procedure. The lower ArT and XIT show that the auto-positioning is more efficient than conventional patient (re)positioning. Notably, none of the auto-positioning runs require any X-ray imaging for manual fine-tuning; hence the patient did not receive any radiation (\ie, XID $= 0$) before the actual CBCT scan. In Table~\ref{table:3Dkeypoints} shows that our method can successfully move the C-arm from head to target regions in one-pass (1 OSR). 
The clinical validity is reported in Table~\ref{table:3Dkeypoints}, bottom.

\begin{figure}
\centering
\begin{subfigure}{1\textwidth}
    \includegraphics[width=1.\textwidth]{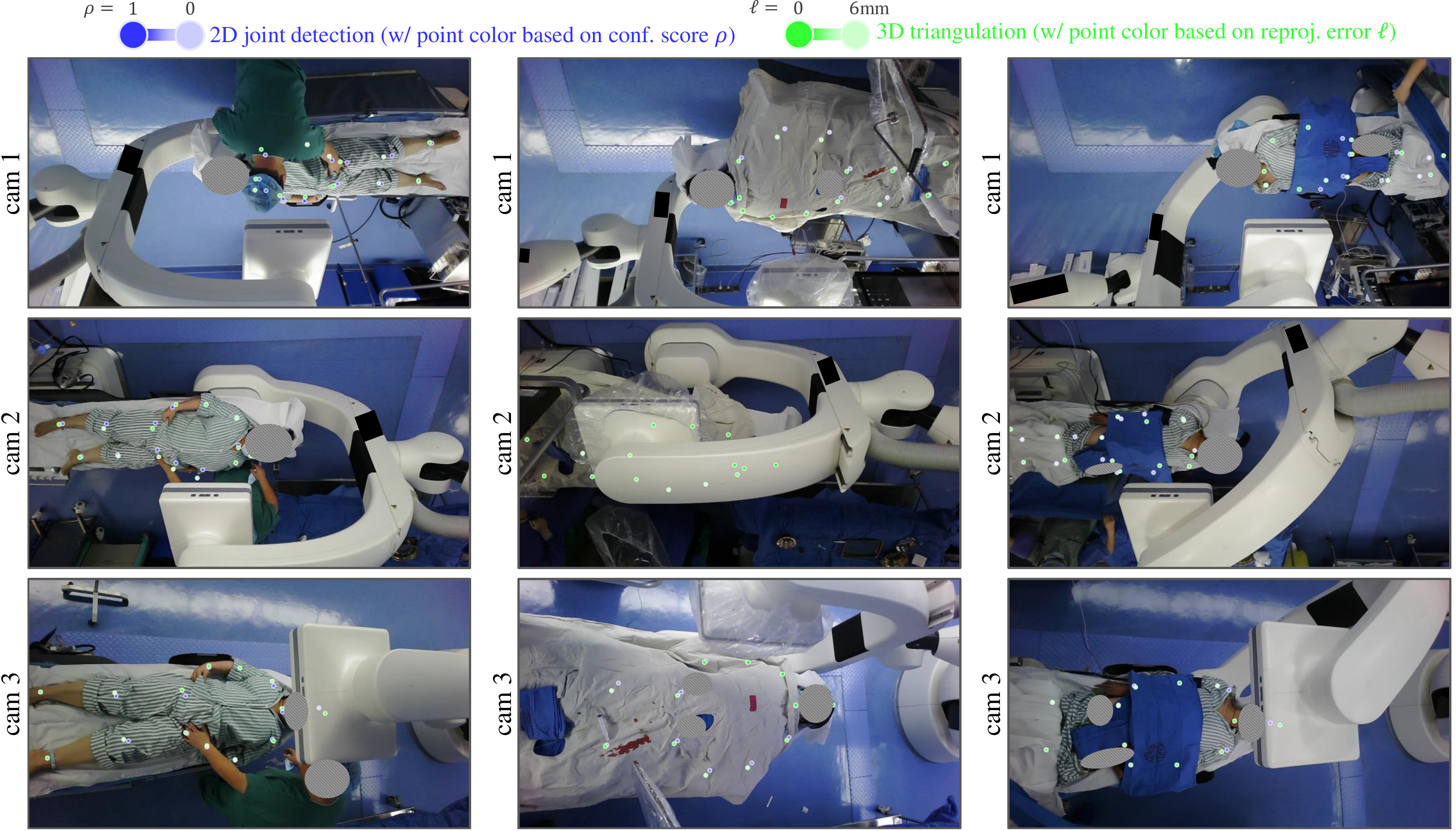}
    \caption{Qualitative results \wrt 2D detection and 3D triangulation of joints. Note that we redacted the images (grayed ellipses) to protect privacy and hide bloodied body parts.}
    \label{fig:qual_2d3d}
\end{subfigure}

\begin{subfigure}{1\textwidth}
    \includegraphics[width=1\textwidth]{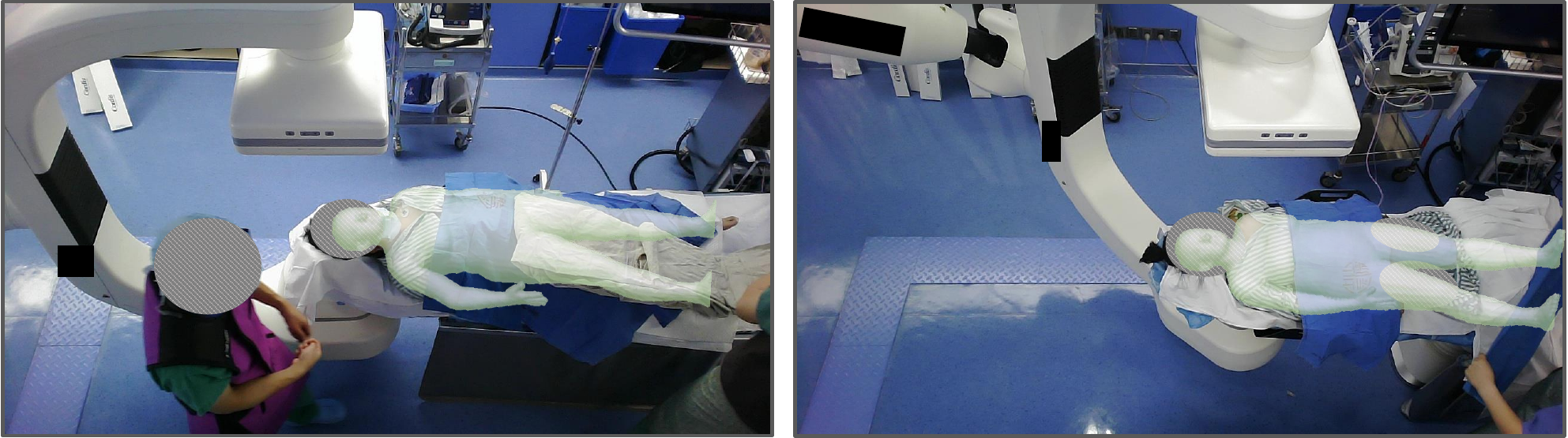}
    \caption{Qualitative results \wrt regression of 3D patient body mesh.}
    \label{fig:qual_mesh}
\end{subfigure}      
\caption{Qualitative results from clinical evaluations.}
\label{fig:qualitative}
\end{figure}

Fig. \ref{fig:qual_2d3d} and \ref{fig:qual_mesh} provide some qualitative results \wrt different components of the 3D patient modeling system.
Fig. \ref{fig:qual_2d3d} presents prediction results of the proposed 2D body keypoint detection model (blue dots) over triplets of synchronized images from the multi-view sensor setup (\ie, \texttt{cam 1}, \texttt{cam 2}, \texttt{cam 3}), as well as the corresponding 3D triangulation results, reprojected into their respective 2D screen space (green dots). The shades of blue and green encode respectively the confidence score $\rho$ returned by the detection model for each prediction, and the mean reprojection error for each triangulated body joint. 
\Ie, the higher $\rho$ is, the more confident the model is in its prediction (and the darker the corresponding blue dot is depicted). Visual ambiguities and occlusions typically result in lower $\rho$. This signal is then leveraged during our custom two-phase triangulation to weight the contribution of each view-specific 2D keypoints to the 3D output (predictions with lower confidence will be given less consideration), as well as during temporal consolidation (see \textit{Method} section for details).
Similarly, a notion of pseudo confidence \wrt triangulation results can be obtained through the computation of the mean reprojection error $\ell$, \ie, the mean $\ell_2$ distance between provided 2D keypoint positions and corresponding 3D predictions reprojected into image space. In Fig. \ref{fig:qual_2d3d}, the higher the reprojection error is, the lighter the shade of green is for the corresponding dot. Similar to $\rho$, the reprojection error $\ell$ is used downstream as a signal to discriminate triangulation results obtained at different time steps, further improving the system's robustness to transient occlusions (\eg, C-arm or medical staff temporarily occluding the patient for one of the cameras), as shown in Fig. \ref{fig:qual_2d3d}. We can indeed observe how the proposed system is robust to heavy patient occlusions (\eg, from C-arm, clinical staff, surgical sheets).

In Fig. \ref{fig:qual_mesh}, some original camera frames are overlaid with their corresponding predicted 3D patient body mesh, computed by the proposed regression model from the triangulated keypoints, and rendered back into a 2D image using the camera extrinsic and intrinsic parameters (note that the parametric human mesh used by our system does not model facial features and expressions, which is why we do not need to anonymize the mesh face).
We can observe how the predicted full-body meshes properly align to the corresponding 2D representations of the patients, matching their pose and overall body shape.
We can however note that our system is not always as accurate for lower body parts, \eg, feet. For generalizability purposes, the deep-learning mesh-regression algorithm used here is trained on pairs of triplet joints (input) and mesh parameters (output) that are synthetically generated from a custom distribution extracted from public datasets. However, most public datasets only contain pictures and annotations for people standing (as opposed to lying down). This training distribution bias explains the regression error \wrt  feet positioning.
This bias impacting lower body parts has not been an issue during clinical trial since tests and actual interventions are targeting top body parts (\eg, head and wrist). Refining the mesh pose distribution and retraining our system accordingly is nevertheless one of our priority for future work.

\begin{figure*}[t]
	\centering
	\includegraphics[width=1\textwidth]{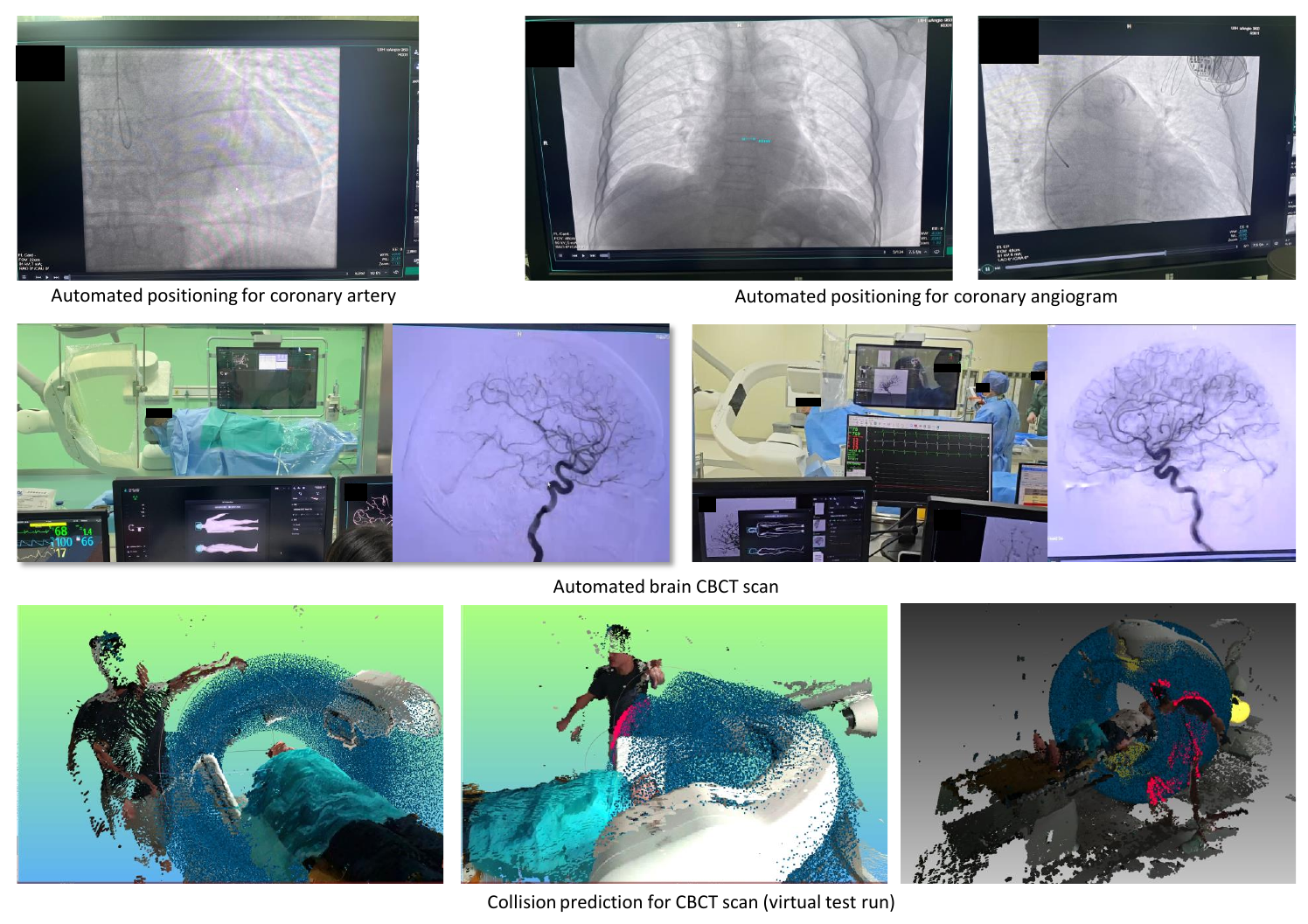}
	\caption{Qualitative results of \textit{AutoCBCT} and \textit{VTR} for collision detection (highlighted red regions).
	}
	\label{fig:vis}
\end{figure*}

\begin{table}[!t]
\small
\center
\scalebox{0.95}{
\begin{tabular}{c|c|c|c|c|c}
\toprule
                  
                  \textbf{Lab RF} & ArT(s)$\downarrow$ &ITs$\downarrow$ & OSR$\uparrow$&XIT (s)$\downarrow$&XID (mGy)$\downarrow$\\
                  \midrule
                  Conventional&  31& 7& 0& 7.85& 2 \\  
                  Manual&  41 & 9& 0&8.05& 2 \\  
                  \textbf{Auto}& \textbf{10} &\textbf{2} &\textbf{100}\% &\textbf{0} &\textbf{0}  \\ 
                  \bottomrule
\end{tabular}}
\scalebox{0.95}{
\begin{tabular}{c|c|c|c|c|c}
\toprule
                  
                  \textbf{Lab RW} & ArT (s)$\downarrow$ &ITs$\downarrow$ & OSR$\uparrow$&XIT (s)$\downarrow$&XID (mGy)$\downarrow$\\ \midrule
                  Conventional&  40 & 8& 0& 9.53& 2 \\  
                  Manual&  37 & 9& 0& 8.66& 2 \\  
                  \textbf{Auto}&\textbf{10}  &\textbf{2} & \textbf{100}\%& \textbf{0}&\textbf{0}  \\ \bottomrule
\end{tabular}}
\scalebox{0.975}{
\begin{tabular}{c|c|c|c|c|c}
\toprule
                  
                  \textbf{Clinical RW} & ArT (s) &ITs & OSR&XIT (s)&XID (mGy)\\ \midrule
                  Auto &  9.24& 2& 100\%& 0&0  \\ \bottomrule
\end{tabular}
}
\caption{Comparison of patient-positioning solutions in lab and clinical settings. ``RF" and ``RW" denote right femoral artery and right radial artery.}  
\label{table:3Dkeypoints}
\end{table}

\begin{table}[!t]
\small
\center
\scalebox{1}{
\begin{tabular}{c|c|c}
\toprule
                  \textbf{Lab} & CDP & TRT (s)\\ \midrule
                        Manual& 100\%& 16 \\
                        \textbf{VTR}& 100\%& \textbf{2} \\\bottomrule
\end{tabular}}
\scalebox{1}{
\begin{tabular}{c|c|c}
\toprule
                  
                  \textbf{Clinical} & CDP & TRT (s)\\ \midrule
                        Manual& 100\%& 13 \\
                        \textbf{VTR}& 100\%&\textbf{2}  \\\bottomrule
\end{tabular}}

\caption{Left: Results on 3D keypoint-estimation-based auto patient positioning  in clinical setting. Right: Results on \textit{VTR} in both lab and clinical settings. }  
\label{table: 3dkpC&vtr}
\end{table}

\paragraph{Virtual Test-Run Module}
In both lab and clinical environments, the proposed \textit{VTR} can achieve 100$\%$ CDP with much lower time (Table~\ref{table: 3dkpC&vtr} right).  The qualitative results of \textit{VTR} are showed in Fig.~\ref{fig:vis}, highlighting how the proposed method can effectively localize obstacles in heavily cluttered environments. There are factors which may have an impact on the \textit{VTR} performance such as the material of objects \wrt camera characteristics.  

\section{Discussion}

Having completed one stage of multi-site clinical trial, \textit{AutoCBCT} is the first vision-based fully-automated C-arm CBCT system validated by clinical trial, to the best of our knowledge. 

Systems that are closest to ours--\eg, targeting conventional CT \cite{karanam2020towards} or gantry-based LINAC \cite{loy2018context} systems---only provide visual feedback, not meeting accuracy requirements and lacking system integration to act upon their predictions. Furthermore, by targetting the interventional environment and C-arm CBCT, our system faces much higher complexity (higher degree of freedom \wrt robot, bed, patient) and clutter (\cf clinical staff,  equipment,  drapes, \etc). We tackle these via exhaustive workflow coverage, \ie, from calibration, to 3D reasoning, to C-arm positioning; whereas prior art is limited to specific steps.

As shown in the clinical results shared in Subsection \ref{sec:clinical_val}, the proposed system improve the efficiency and throughput via accurate automation, replacing joystick controls that are unintuitive even for experienced physicians. We believe that user experience can be further improved by leveraging the proposed system, as it can enable the development of smarter, simplified workflows and user interfaces to help flatten learning curve for novice operators and minimize clinicians' cognitive disruption.

Our evaluation shows that automated positioning can be more accurate than manual workflow, as highlighted for example by the perfect score \wrt one-pass successful rate (OSR) in Table \ref{table:AutoCBCT}. 
Thanks to our contributions, the accuracy \wrt scanner positioning for radial-artery imaging is also much below the requirements (37mm) fixed after consultation with experts. As deep-learning models keep improving and relevant data is made more and more available, we are convinced that the overall accuracy can increase significantly in the coming years, enabling stricter applications (\eg, radiotherapy guidance).

Automation also means more consistent scanning procedures, which can facilitate downstream analysis of the imaging results, as well as enable patient-specific protocols, \eg, adapting the C-arm path (collision avoidance) or X-ray dose according to the patient's body dimensions, an increasingly important feature as the bariatric population increases.
Last and not least, our goal with \textit{AutoCBCT} is also to make the procedure safer not only for the patients (as already highlighted via the lower X-ray exposure presented in Table \ref{table:AutoCBCT}), but also for the clinicians in the operating room.
\Ie, down the line, \textit{AutoCBCT} could also unlock new applications, such as the real-time tracking of radiation exposure to clinicians. 

Our system still faces some technical limitations, such as its handling of transparent and dynamic objects during test runs, \eg, due to intrinsic limitations of the sensor setup. Moreover, while our system shows great accuracy for apparent body parts, it inherently faces more challenges for the positioning \wrt inner body structures, \eg, organs. Organ targeting must rely on a statistical modeling of the patient anatomy conditioned on their body shape, which may not fit all populations. Recent works \cite{guo2022smpl,liu2023implicit} have been focusing on improving the modeling of anatomical structures based on visual information, leveraging auxiliary information such as preoperative scans of the patient. It is a exciting but challenging research direction, and many obstacles must still be faced towards registering pre- and intra-operative modalities.

\section{Methods}
\label{sec:met}

\begin{figure*}[t]
	\centering
	\includegraphics[width=1\textwidth]{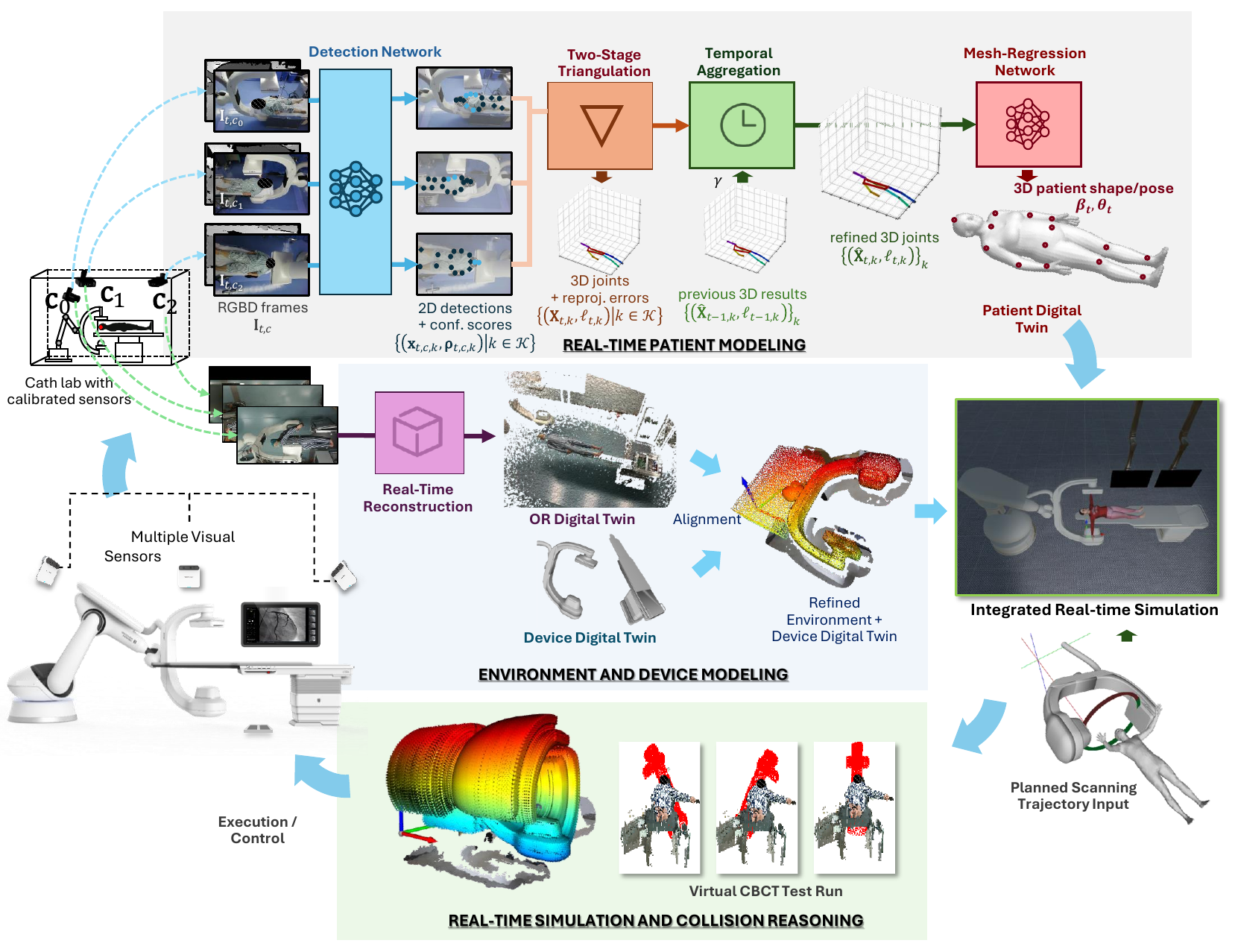}
	\caption{System overview. (A) Proposed 3D patient body modeling pipeline, refining the patient's pose and shape at each timestep. 
	(B) Proposed virtual test run pipeline, detecting possible collisions during scanning.}
	\label{fig:3dhpe_vtr}
\end{figure*}

\noindent
\subsection{Overview}
The proposed vision-based C-arm system (Fig.~\ref{fig:3dhpe_vtr}) is comprised of three major components: (1) a triple-view RGBD camera system, (2) a multi-view 3D patient body modeling, and (3) a virtual test run.
The RGBD camera system captures the environment information, \ie, RGB and depth images, from three different views at each time step (Fig.~\ref{fig:3dhpe_vtr} A). Using the multi-view RGBD frames, the proposed 3D patient body modeling method estimates the 3D pose and shape of the patient, and the spatial location of target region.
The target region is selected by the operator on a given general human body model shown on a monitor. Receiving the estimated target location, the C-arm robot can then automatically move to the target.
Once arrived at the specified location, the virtual test run (Fig.~\ref{fig:c-arm_overview} B) is performed to detect any obstacles in the rotation path.

\noindent
\subsection{Multi-view RGBD Setup}
The goal of the multi-view RGBD camera system is to capture 3D scene information from different viewpoints (3 in our system), each providing color and depth (2.5D) information (Fig.~\ref{fig:c-arm_overview} B). 
To mitigate the impact of occlusion, the position of the three cameras are configured as roughly shown in Fig.~\ref{fig:c-arm_overview} A. 
To ensure the 3D scene information is useful to the C-arm robot system, camera calibration is performed to compute the extrinsic parameters between each camera and the robot coordinate system (\ie, camera extrinsics). The three RGB cameras $\mathcal{C}$ are calibrated by leveraging markers preplaced on the DSA system (support)  and the calibration workflow from \cite{karanam2020towards}. Here, we use one camera $c \in \mathcal{C}$ as an example to demonstrate the workflow. Given a marker, $p$, in C-arm robot system, its 3D coordinate is $\mathbf{P}=[x, y, z]$. The corresponding undistorted projection of the $\mathbf{P}$ on the image plane of the camera $c$ is $\mathbf{p}_c=[u, v, 1]$. The rotation matrix and translation vector of the camera are represented by $\mathbf{R}_c$ and $\mathbf{t}_c$, respectively. Thus, we have $\mathbf{p}_c=\mathbf{K}_c(\mathbf{R}_c \mathbf{P}+\mathbf{t}_c)$,
where $\mathbf{K}_c$ is the intrinsic matrix, determined in an offline process using a standard
checkerboard target~\cite{zhang2000flexible} and remains unchanged after system
installation.  To compute each $\mathbf{R}_c$ and $\mathbf{t}_c$, we establish 2D-3D correspondences of multiple marker over various vertical and horizontal positions. Formally, given the 2D-3D correspondences set $(\mathbf{M}_i$, $\mathbf{m}_{i,c})$, where $\mathbf{M}_i$ is a set of 3D locations of $i$-th marker and $\mathbf{m}_{i,c}$ is a set of its 2D image projections. This set is then used to solve the $\mathbf{R}_c$ and $\mathbf{t}_c$ 
using standard robust perspective-n-point solvers~\cite{lepetit2009epnp}. Once the
calibration process is successfully conducted for three cameras, 3D point clouds can be generated from depth images captured from any view by using the estimated intrinsic and extrinsic parameters. 
Finally, fusing the these point clouds together yields a complete point cloud, describing the 3D representation of the operating room (Fig.~\ref{fig:c-arm_overview} C).      



\noindent
\subsection{Multi-view 3D Patient Body Modeling}
Fig.~\ref{fig:3dhpe_vtr} (A) presents an overview of the proposed 3D patient modeling module. Briefly, to infer the 3D pose and shape (resp. $\bm{\theta}$ and $\bm{\beta}$ of SMPL~\cite{loper2015smpl} mesh model) of the patient, the RGBD images from each camera view $c \in \mathcal{C}$ at time $t$ are first passed to a 2D keypoint detection module to infer the 2D locations (in image space) of a set of predefined body joints (noted $\mathcal{K}$), \ie, $\{\mathbf{x}_{t,c,k} \in \mathbb{R}^2 | k \in \mathcal{K} \}$. Given the inferred 2D joint locations from multiple camera views, a custom triangulation is performed~\cite{hartley2003multiple} to infer the joint locations in 3D space, based on the cameras' intrinsic and extrinsic parameters. These 3D keypoints are then refined via temporal consolidation to account for environmental challenges (such as occlusions) and are finally fed into a synthetically trained CNN model \cite{qi2016pointnet,3DPatientModel_miccai22} for SMPL pose and shape parameter inference.

\subsubsection{2D Keypoint Detection}
Specifically, the RGBD images are first forwarded to a 2D CNN-based keypoint detector to infer a collection of $k$ predefined body joint positions in each image space. 
Keypoint detection is typically achieved by utilizing deep convolutional neural network (CNN) models in recent computer vision studies. 
Here we adopt HRNet~\cite{sun2019hrpose} as the foundational architecture and utilize a custom dual-stream RGB-D structures to enhance corrupted RGB features from depth images \cite{3DPatientModel_miccai22} to mitigate occlusion problem caused by fully-/partially-covered patient. 
Most recent works in 2D pose estimation~\cite{openpose_pami,he2017mask,sun2019hrpose} are essentially single-modality (\ie, RGB-only) solutions learned from open-source in-the-wild human pose data \cite{coco2014,human36m}, with most of person subjects standing upright performing daily activities. They may easily fail in specialized situations when applying in real-world clinical applications where patients are usually covered by a cloth or surgical sheet, or occluded by medical equipment or devices.
We thus propose a multi-modality 2D keypoint detection framework that leverages cross-modality complementary sensory information, applicable to generic as well as specialized clinical scenarios (\eg, patients lying under surgical sheets) with cluttered environments.


\begin{figure*}[t]
	\centering
 	\includegraphics[width=1\linewidth]{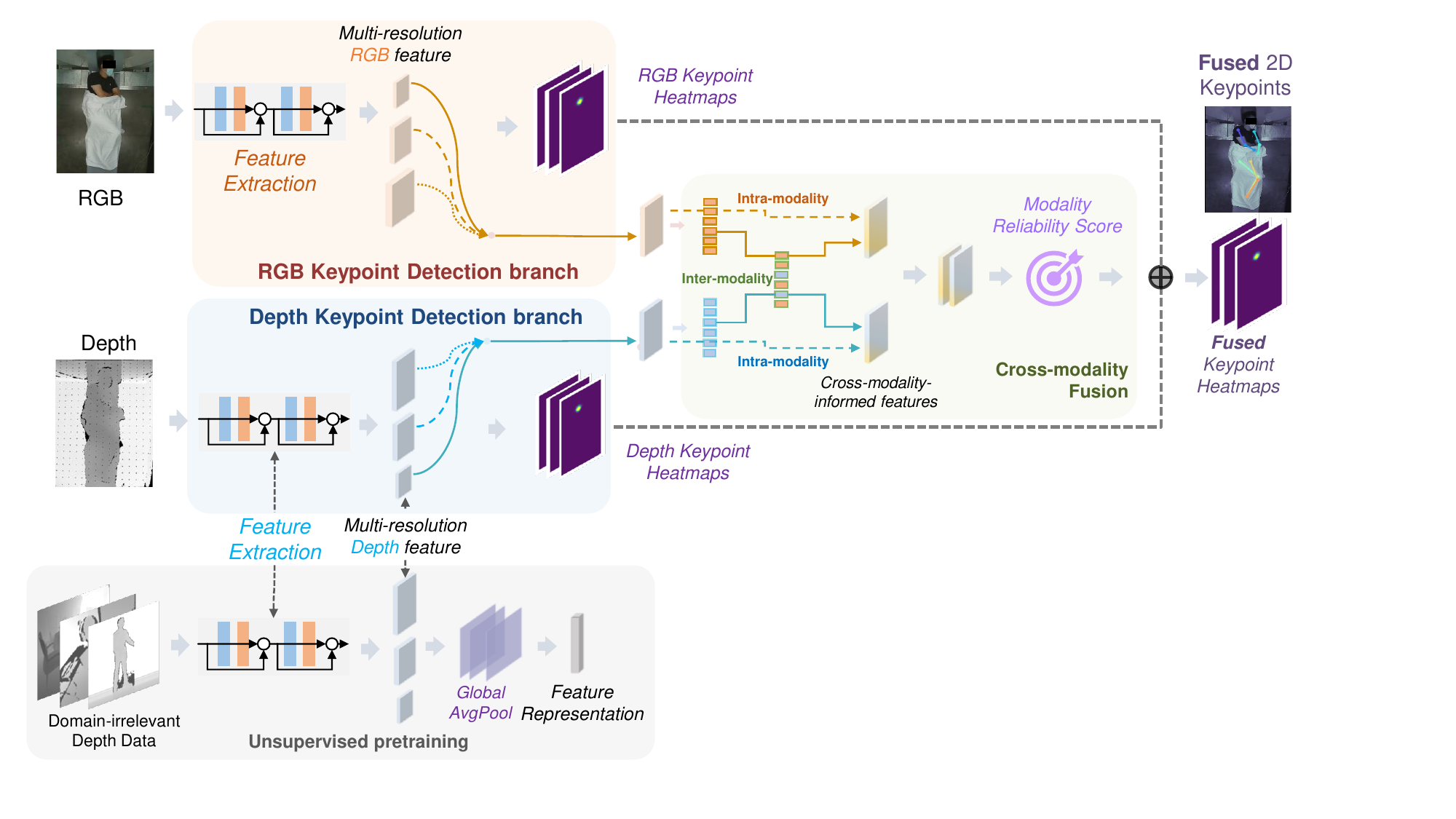}
	\caption{Proposed RGBD keypoint detection framework with attention fusion.}
	\label{fig:fusion}
\end{figure*}

While RGB-based keypoint detection network can be learned from large-scale open-source human-pose-estimation datasets \cite{coco2014,human36m,mpii_16cvpr,LSP_Johnson10}, the generalizability of such a single-modality keypoint detection network may be easily compromised in real clinical environments with significant domain gap and heavy occlusions. We thus propose to leverage an additional sensor modality, \ie, depth data captured by sensors measuring the distance to surfaces in its field of view, in order to distill complementary patient body pose/shape information and thus to minimize the impact of patient covers and clothes. As the learning of depth-based keypoint detector may easily suffer from poor generazability due to the lack of large-scale depth-based human pose estimation datasets, we propose to perform a self-supervised learning scheme \cite{chen2020simsiam} to first learn generic depth-image representations from domain-irrelevant unlabeled data, and then finetune with domain-relevant samples with person-pose annotations, to improve the robustness and accuracy of the depth-based keypoint detector. 

Multi-modality (RGB and depth) images offer complementary insights into the patient's body pose and shape information, with each modality presenting unique advantages across different scenarios. For instance, when a surgical sheet obscures the patient (Fig.~\ref{fig:fusion}), RGB features are significantly impacted by occlusion, while depth data still retains valuable shape and contour details beneficial for modeling the patient's body.
Our objective is to devise a cross-modality fusion network capable of effectively integrating the complementary information from RGB and depth images through intra- and inter-modality feature aggregation, thereby enhancing keypoint detection performance. To achieve this, we propose a two-stream reliability score-based RGBD fusion network, as depicted in Fig.~\ref{fig:fusion}. This fusion network leverages the final stage features from the RGB and depth branches of the HRNet backbone, routing them through a fusion module equipped with intra-modal and inter-modal feature aggregation for reliability score prediction. Specifically, we evaluate the reliability of RGB-based and depth-based single-modality keypoint detection by predicting a score (range from 0 to 1) through a binary classifier, given the predicted single-modality joint heatmaps. The binary classifier is supervised by the reliability labels, \ie, whether RGB or depth modality is giving the most accurate prediction results given the prediction and ground-truth difference.
This score is then utilized to weight the keypoint heatmap predictions from each modality to generate the fused cross-modality keypoint heatmaps. Through this mechanism, the proposed module seamlessly integrates complementary information from individual modalities, facilitating the acquisition of enriched feature representations for more precise and resilient keypoint detection.

While top-down 2D keypoint detection networks \cite{sun2019hrpose,he2017mask} assume the input image always contain the full person body (\ie, the complete set of predefined body joints is visible in the image space), false positives can easily be introduced when the target person is not fully covered by the camera frustum. We thus introduce a \textit{visibility} auxiliary network which takes the predicted $k$ fused keypoint heatmaps as input and output $k$ visibility scores $v_k$ corresponding to each joint, to better avoid false alarms \wrt keypoint detection.
The input $k$ keypoint heatmaps are sent through $N$ layers (we set $N=5$ in our current system) of 2D convolutional layers and a binary classifier after heatmap flattening to predict whether a joint is visible ($v_k=1$) or not ($v_k=0$). 
On the other hand, as keypoint detection can also be incorrect or noisy due to the extremely challenging clinical setting, we further introduce a \textit{confidence} auxiliary network to predict $k$ confidence scores $\rho_k$ respective to each 2D predictions, supervised by the normalized $\ell_2$ distance between predicted and ground-truth joint locations, to reflect the reliability and correctness of our 2D keypoint prediction workflow. The confidence auxiliary network has similar architecture as the visibility auxiliary network, with the last layer replaced by a linear regressor followed by a softmax operator to predict the confidence score range from 0 to 1.

It should further be noted that some body keypoints have higher clinical value. For example, digital subtraction angiography is often used during procedures focused on the patient's radial artery (\eg, for transradial cardiac catheterization) or head (\eg, intraoperative CBCT scan for brain aneurysm treatment), thus justifying higher precision requirements for the positioning of the corresponding body keypoints (head center, left/right radial artery).
Therefore, for the 2D detection of these keypoints, we adopt a second-stage refinement procedure based on specialized detection networks.
Our solution crops an image patch for each region of interest, based on the main set of detection results (\eg, using the predicted wrist and elbow positions to crop an image patch containing the patient's hand and forearm), then passes this image patch to a detection neural network. Each specialized detection network is using the same architecture and training regimen as the full-body detection model, but is optimized for 2D keypoints specific to the target body region (\eg, hand keypoints or facial keypoints).
If a region-specific detection succeeds, its results are used to overwrite the corresponding target 2D keypoint returned by the system. 

\subsubsection{3D Keypoint Triangulation}
The proposed two-stage confidence-weighted triangulation takes as input the sets of predicted 2D keypoints and confidence scores \wrt each camera $c$, and returns a set of keypoints $\{\mathbf{X}_{t,k} \in \mathbb{R}^3 \mid k \in \mathcal{K} \}$ in the C-arm robot's 3D coordinate system, computed as follows:
\begin{equation} 
\label{eq:triang}
\begin{aligned}
    \mathbf{X}_{t,k} &= \argmin_{\mathbf{X} \in \mathcal{X}_{t,k}}\sum_{c \in \mathcal{C}}{\mathbf{\rho}_{k,c,t}\ell_c(\mathbf{x}_{t,c,k}, \mathbf{X})} \\
    \text{with } \mathcal{X}_{t,k} &= \{\argmin_{\mathbf{X}\in\mathbb{R}^3}\sum_{c \in C}{\ell_c(\mathbf{x}_{t,c,k}, \mathbf{X})} \mid C \subseteq \mathcal{C} \text{ with } |C| \geq 2 \}, \\
\end{aligned}
\end{equation}
where $C$ is subset of at least two cameras, and $\ell_c(\mathbf{x}, \mathbf{X}) = \|\mathbf{K}_c(\mathbf{R}_c\mathbf{X} + \mathbf{t}_c) - \mathbf{x}\|_2$ is the reprojection error \wrt camera $c$.
During the first stage, for each combination $C$ of 2+ cameras, we triangulate each joint $k$ according to the 2D predictions over the corresponding frames, \ie, using singular-value decomposition (SVD) to obtain the 3D candidate which minimizes the mean reprojection error $\sum_{c \in C}{\ell_c(\mathbf{x}_{t,c,k}, \mathbf{X})}$. 
We thus obtain a set $\mathcal{X}_{t,k}$ of 3D candidates for each joint. 
During the second stage,  
the model returns the best candidate, \ie, which minimizes the reprojection error averaged over \textit{all} cameras $\mathcal{C}$, weighted by the respective detection confidence scores.
By considering subsets of views and confidence scores, our two-step triangulation is robust to detection errors (\eg, when the patient is partially occluded in a frame, leading to erroneous joint detection).
Note that for each keypoint, the model also returns a scoring vector $V_{t,k} = \{\rho_{t,k}, v_{t,k}, \frac{1}{\ell_{t,k}}\}$, averaging the per-view scores over the subset of views $C$.

\subsubsection{Temporal Consolidation of 3D Predictions}

To further improve the robustness to transient occlusions and other noise sources, we propose to leverage temporal consistency priors.
\Ie, while the upstream modules return 3D candidates at each time steps, said candidates can be noisy or even erroneous (due to occlusions, sensing errors, algorithmic uncertainty, \etc). 
The main challenge thus consists of accurately discriminating between noise and actual motion, when consecutive keypoint predictions are not consistent. If the temporal discrepancy is caused by actual motion of the patient's body, the updated position should be returned. If instead the temporal discrepancy is caused by prediction uncertainty, the system should return the most confident results within a window of past predictions.
This quality assessment and consolidation task is made even more difficult by the open-set nature of the noise sources, preventing the adoption of data-driven mitigation measures. 


\begin{figure*}[t]
	\centering
	\includegraphics[width=1\textwidth]{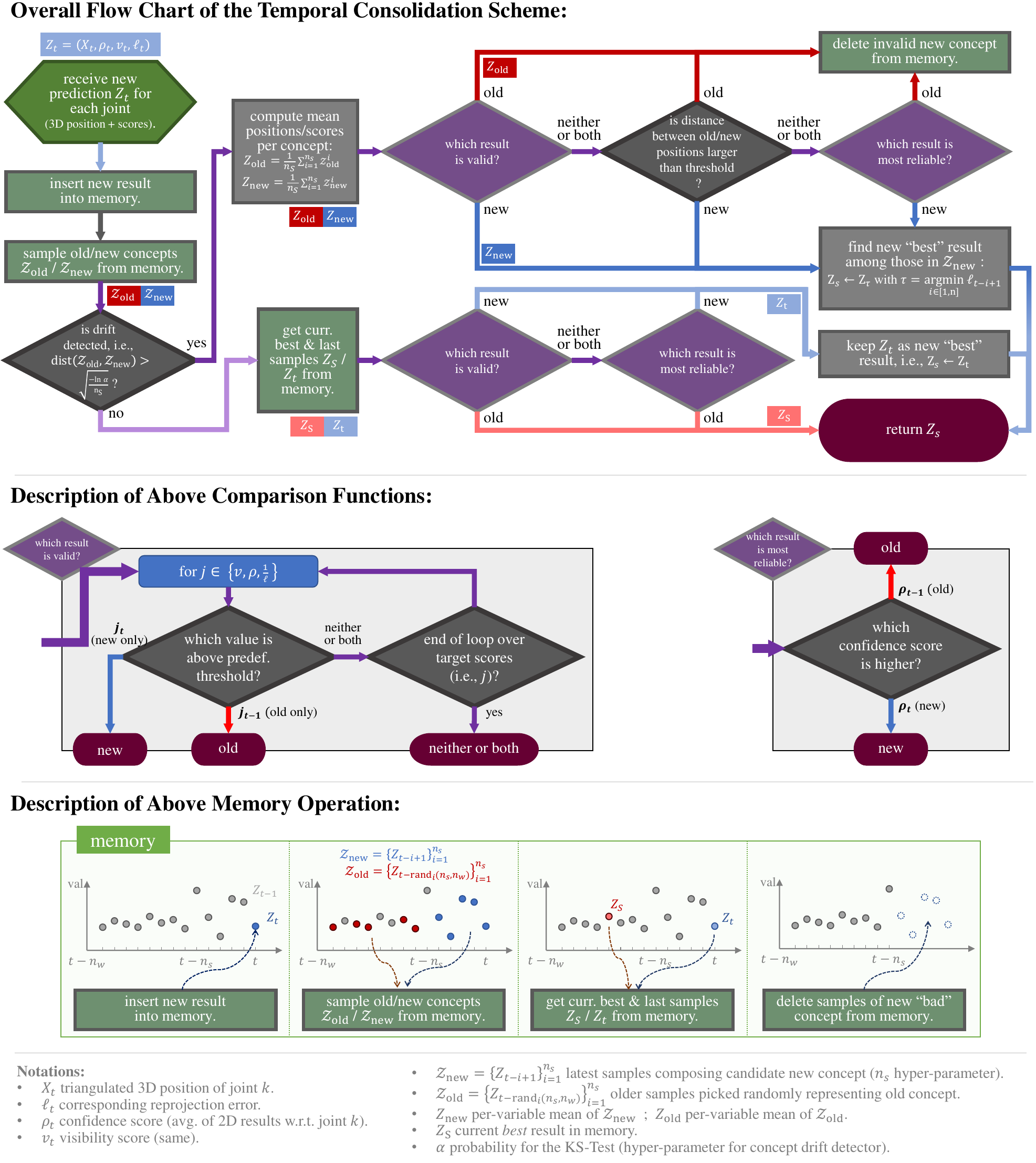}
	\caption{Flow chart detailing the temporal analysis of 3D keypoint predictions.}
	\label{fig:temporal_workflow}
\end{figure*}

Therefore, we propose the following temporal-aggregation module. At each time step $t$ and for each keypoint $k$, the module receives the latest scored candidate $Z_{t,k} = (X_{t,k}, V_{t,k})$ (we later drop the subscript $k$ for clarity), memorizes it, and returns a more consistent result $Z_{s,k}=(X_{s,k},V_{s,k})$ that accounts for previous results $\{Z_{t-1}, Z_{t-2}, ..., Z_{t-w}\}$ (with $w$ its memory size).
Our module relies on multi-variate unsupervised online concept-drift detection algorithms \cite{dos2016fast,wang2018systematic,gemaque2020overview} to analyze any change in the temporal distribution of per-keypoint predictions.
As drift detector, we adopt a version of the Kolmogorov-Smirnov (KS) windowing scheme \cite{dos2016fast,zhao2017kolmogorov,togbe2021anomalies}, that leverages two-sample two-sided Kolmorogov-Smirnov statistics \cite{massey1951kolmogorov,fasano1987multidimensional,berger2014kolmogorov} to compare distributions, adapted for multivariate data. 
Inspired by \cite{rabanser2019failing}, our per-keypoint drift detectors applies the KS test on the marginals of the features, for each feature dimension (3D coordinates and score), and returns the maximum score. Bonferroni correction and stable recursion \cite{viehmann2021numerically} are employed to compute the p-values, used to generate the binary outcome (``is the distribution drifting?'').

If a change in the distribution of $Z$ is detected (\ie, if a drift in the consecutive values of $X$ or $V$ is detected, \eg, confidence drift and/or localization jitter due to increasing patient occlusion), the processor partitions the past results into at least 2 subsets (“concepts”); one subset corresponding to the results before the drift, and one subset corresponding to the results after. A set of score-based conditions are then applied to figure out which distribution can be deemed \textit{reliable}, \ie, with scores above some predefined thresholds (\eg, if some body part becomes occluded, new localization results would become unreliable, which should translate into having one or more corresponding scores in the new $V$ distribution drop significantly).
If one distribution (old or new) is deemed valid but not the other, the system adopts the valid concept. If both are valid or invalid, a second set of score- and distance-based conditions is applied to select the output concept, \ie, to decide if motion should be considered (above-threshold distance between $X$ distributions) or to decide which concept is more accurate.

Once the model selects which concept to keep, the data points within the corresponding subset are compared and the one deemed most reliable (based on scores $V$) is finally returned as $Z_s$. All the aforementioned conditions are detailed in the flow diagram presented in Fig.~\ref{fig:temporal_workflow}.

\subsubsection{3D Patient Body Mesh Regression}
To recover the SMPL mesh representation~\cite{loper2015smpl} of the patient based on the estimated 3D keypoints $\{\mathbf{\hat{X}}_{t,k} \in \mathbb{R}^3 \mid k \in \mathcal{K} \}$, we deploy a PointNet-based model \cite{qi2016pointnet} which takes sets of 3D keypoints as input and outputs corresponding SMPL $\bm{\theta}$, $\bm{\beta}$ parameters. 

Specifically, to learn the PointNet model, we generate synthetic data pairs (SMPL $\bm{\theta}$, $\bm{\beta}$ and corresponding 3D keypoint locations inferred from rendered mesh given $\bm{\theta}$, $\bm{\beta}$) since SMPL annotations are typically extremely challenging to be directly generated from person images or 2D/3D keypoint locations \cite{loper2014mosh}. 

After regressing the SMPL pose and shape parameters from the PointNet-based model, the system is able to generate a 3D mesh representation of the patient, properly registered in the 3D coordinate system of the operating room (see qualitative results in Fig.~\ref{fig:qual_mesh}). 
Formally, SMPL provides a differentiable kinematic function $\mathcal{S}$ from $\{\theta, \beta \}$  to a 3D mesh of 6,890 vertices: $v=\mathcal{S}(\theta, \beta) \in \mathbb{R}^{6,890 \times 3}$. 
As a result, the 3D joint location of the $\mathcal{K}$ joints of interest can also be regressed as  $\widetilde{\mathbf{X}}_t =\mathcal{J} \cdot v$, with $\mathcal{J} \in \mathbb{R}^{k \times 6,890}$ a pre-trained linear regression matrix.

\subsection{Virtual Test Run}
The virtual test run (\textit{VTR}) is defined to replace manual test runs, so as to efficiently detect obstacles. The \textit{VTR} pipeline is briefly depicted in Fig.~\ref{fig:3dhpe_vtr} (B). Specifically, a fused point-cloud of the operating room, $\mathbf{P_{or}}$, is first generated by using the calibrated depth images from three views in robot coordinate system. A second point-cloud representing the C-arm, $\mathbf{P_{ca}}$, is generated by sampling points over the surface of the C-arm CAD model.
According to the motion parameters from the C-arm robot system, $\mathbf{P_{ca}}$ and $\mathbf{P_{or}}$ can be aligned (middle of Fig.~\ref{fig:3dhpe_vtr} (B) ), and the C-arm in $\mathbf{P_{or}}$ can be subtracted, resulting in a point-cloud $\mathbf{P^{new}_{or}}$ only capturing potential obstacles. 
Based on the C-arm robot's parameters, its movement pattern can be simulated with $\mathbf{P_{ca}}$ before CBCT. 
Obstacles can be found by conducting the simulation within $\mathbf{P^{new}_{or}}$. 
This way, the test run task of detecting obstacles can be converted into finding collision regions between two point clouds, \ie, $\mathbf{P_{ca}}$ and $\mathbf{P^{new}_{or}}$. 
Following the algorithm proposed in \cite{klein2004point}, we can efficiently detect if the point clouds collide and where collided regions are, consequently obtaining the 3D locations of actual obstacles. This virtual means of detecting obstacles is much more efficient than conventionally manual approach.

Note that, as mentioned in our \textit{Discussion}, a current limitation of the system is the handling of translucent objects.
Due to the reflection and refraction properties of light and the characteristics of depth camera, transparent object surfaces are not captured in depth images and thus are not represented in the point-cloud used for simulation. Collision detection for transparent and mirroring objects is an ongoing topic of research, which vast real-world applications \cite{yasin2020unmanned,ichnowski2021dex,zhang2021trans4trans}.

\section{Conclusion}
To the best of our knowledge, we proposed the first clinical-ready vision-based fully automated C-arm CBCT system. 
Leveraging RGBD data from multi-view 3D cameras, the proposed system models the patient body and the surgical environment in 3D, and integrates them into a simulation environment to  perform virtual test runs in order to validate planned scanning path and identify potential collisions and obstacles. 
The proposed system enables a more streamlined and efficient workflow for C-arm CBCT intraoperative scans in interventional procedures, which simplifies manual operations, saves treatment time, reduces patient's exposure to radiation, and ensures patient and physicians' safety.  
The efficacy and effectiveness of the system and each component are validated by comprehensively designed experiments in both lab and clinical settings. For future studies, we plan to extend the system to enhancing and automating other workflows in the interventional suite and surgical rooms, and improve the system robustness against other complex clinical scenarios.


\bibliography{dsa}

\appendix
\beginsupplement
\newpage
\noindent
\begin{center}
\pdfbookmark[0]{Supplementary Material}{sup_mat}
{\Large\textbf{\centering \MakeUppercase{Supplementary Material}} \vspace{1em}}
\end{center}

\section{Implementation Details}

\subsection{Multi-Sensor Setup}

\begin{figure}[]
	\centering
	\includegraphics[width=1\textwidth]{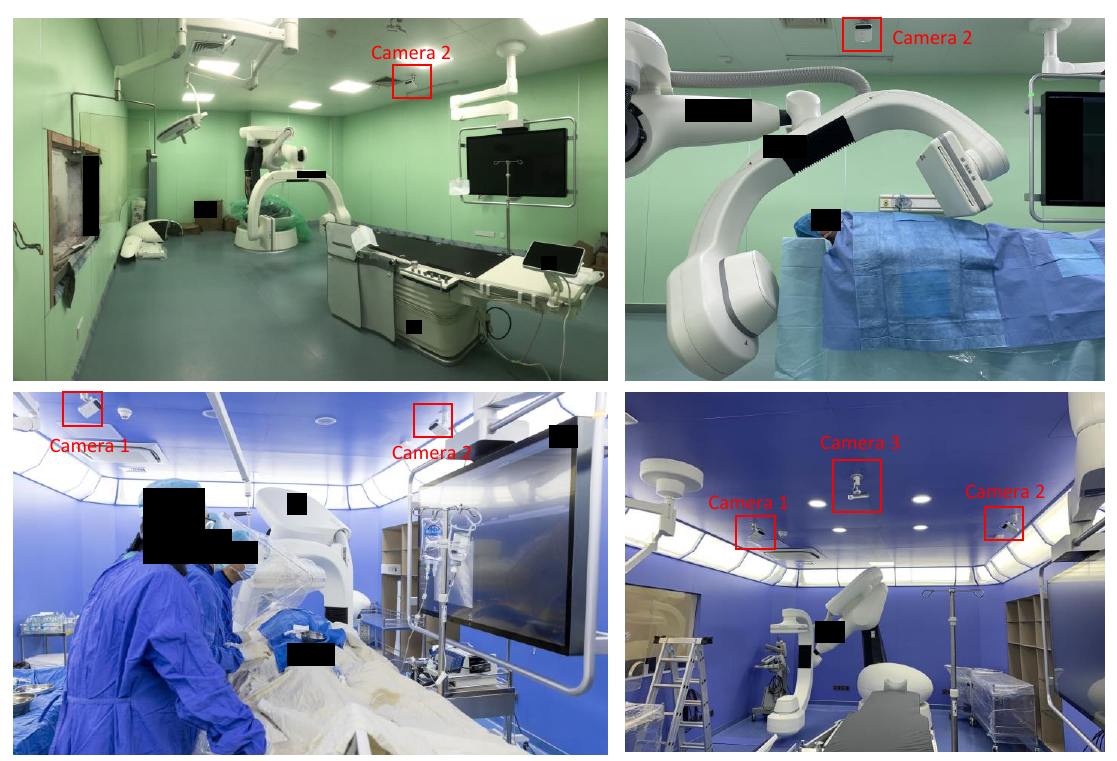}
	\caption{Pictures of the deployed multi-camera systems in angio suites at hospitals.}
	\label{fig:cameras}
\end{figure}

Figure \ref{fig:cameras} shares pictures of the proposed multi-camera system deployed in angio suites at various hospitals, where the proposed system underwent clinical trial.
RGBD sensors currently used are Intel RealSense™ LiDAR Camera L515 ones, though the system is agnostic to camera models and has been tested with others.

\subsection{Patient Positioning Implementation}
As mentioned in the \textit{Methods} section of the main paper, the detection of key body joints $\mathcal{K}$ is performed on each image frame provided by the multi-view sensing system. 
We configure the HRNet model to output a set of 15 keypoints corresponding to predefined body joints $\mathcal{K}$:
\texttt{Right Ankle} (R.Ak.), \texttt{Right Knee} (R.Kn.), \texttt{Right Hip} (R.H), \texttt{Left Hip} (L.H.), \texttt{Left Knee} (L.Kn.), \texttt{Left Ankle} (L.Ak.), \texttt{Right Wrist} (R.Wr.), \texttt{Right Elbow}, (R.El.) \texttt{Right Shoulder} (R.Sh.), \texttt{Left Shoulder} (L.Sh.), \texttt{Left Elbow} (L.El.), \texttt{Left Wrist} (L.Wr.), \texttt{Neck}  (Ne.) \texttt{Head Top} (He.), and \texttt{Nose} (No.).

The auxiliary networks for confidence and visibility scoring share the same shallow CNN architecture, composed of 4 blocks of convolutional and batch-normalization layers with respectively 32 $1\times1$ kernels, 32 $3\times3$ kernels, 128 $1\times1$ kernels, and 256 $1\times1$ kernels; and 1 final fully-connected layer with ReLU activation.
The main HRNet model is first pretrained on public datasets COCO \cite{coco2014}, MPII \cite{mpii_16cvpr}, then fine-tuned in a supervised manner on a smaller set of collected data (both from dummy and clinical settings). Training is done leveraging an Adam optimizer \cite{KingmaB14} with a learning rate of 0.0003. 
The score-regression sub-models are trained jointly, in a self-supervised manner, \ie, enforcing the predicted confidence scores to be proportional to the accuracy of the corresponding keypoint predictions (accuracy computed in an online manner at each training iteration), and enforcing the visibility score to be proportional to the signed distance between the ground-truth 2D keypoint locations and the closest viewport edge.

The keypoint triangulation is performed by iteratively solving the least-square problem defined in Equation 1 of the main paper, applying a standard solver using trust region reflective method and Huber loss \cite{conn2000trust}.
For the temporal change detection, we opt for a window size $w$ of 25 time steps (each sampled every 5 seconds), a statistical size $n_S$ of 10, and we set $\alpha = 0.05$ for the KS test.

For the synthetic training of the PointNet-based 3D mesh regressor, we generate 500k data points with SMPL pose parameter sampled from a distribution learned from public datasets (AMASS \cite{AMASS:2019}, UP-3D \cite{lassner2017unite}, and 3DPW \cite{von2018recovering}), and shape parameter from a Gaussian distribution \cite{sengupta2020synthetic}.
The 3D mesh regression network contains 3 layers of convolution, with each layer followed by batch normalization. Max-pooling operations are performed after convolutional layers to extract global features from input 3D keypoints, and then sent into 3 additional fully-connected layers for final SMPL pose and shape regression. The model is trained for 250 epochs, using an Adam optimizer and a learning rate of $0.0005$. 
Note that the whole solution is implemented using the PyTorch framework~\cite{paszke2019pytorch}.

\subsection{Virtual Test-Run Implementation}
In the virtual test run, we represent the C-arm and the surgical room with two sets of point-clouds, respectively. The collision regions are detected by computing the overlap of the two point-clouds in 3D space. The fine granularity of detection is determined by the resolution of the point-clouds. In our experiments, we crop a 3D space of $3,000\times 2,000 \times 2,000\text{mm}^{3}$ from the observed surgical-room point cloud, corresponding to the volume of the C-arm workspace. This volume is then uniformly downsampled into $100\times 100 \times 100$ points. Therefore, the detection granularity is $30\times 20 \times 20 \text{mm}^{3}$.

\section{Evaluation of Algorithmic Contributions}

The demonstrated clinical value is achieved through the development of a complex vision- and machine-learning-based workflow. 
However, 
state-of-the-art solutions fail to meet the clinical requirements in terms of accuracy and reliability. To ensure the robust functioning of the patient-positioning and test-run modules despite the complexity of target environments, we propose a variety of technical contributions covering every component of the \textit{AutoCBCT} workflow. We first highlight the need for meticulously-crafted functions in the following experiments, before introducing our contributions in more detail in the \textit{Method} section.

\subsection{Experimental Protocol}

\noindent
{\bf Testing Data.}
Relying on the laboratory deployment of the DSA suite, we collect multiple multi-view RGBD image sequences during mock scanning procedures on a human-shaped medical manikin and on actual volunteers. As depicted in 
Fig. 2 of the main paper, the data thus showcase the patient lying on the bed support in supine pose, with and without surgical covers on top of them, and with a variety of additional environmental clutter, \ie, moving C-arm, other persons interactive with the patient, \etc.
We select 6 multi-view sequences and annotate them. For the evaluation of the 2D keypoint detector, we manually annotate 685 frames with the 2D locations of the 15 target body joints, whereas we manually annotate 108 frames with their 3D locations for the evaluation of the 3D mesh-based body part predictions.

To provide an end-to-end assessment of the quality of the 3D positioning, we generate ground-truths  by measuring the positions of the scanning center via laser-line projection, after an expert manually positioned the C-arm according to protocol for the scanning of the corresponding body part. 
Due to the human/machine cost of such a data collection, we focus this technical analysis on one of the most clinically-relevant body keypoints, \ie, the right radial artery. We annotate the corresponding target 3D locations over 892 time steps.

\noindent
{\bf Metrics.}
To thoroughly evaluate our proposed patient positioning workflow, we first evaluate the accuracy of image-based keypoint detection by following the standard protocols for 2D human-pose estimation, \ie, considering the commonly-used 2D mean per joint position error (MPJPE) and percentage of correct keypoints (PCK) \cite{2deval_bm2018} as evaluation metrics. More specifically, we consider the ``percentage of correct keypoints (PCK) -- torso" (distance from shoulder to hip) thresholded at $0.3$ (\cf clinical requirement) \cite{mehta2017monocular}.
With regard to the accuracy of reconstructing a 3D mesh of the patient's body, we compute the mean $\ell_2$ distance between the set of predefined 3D joints regressed from the predicted mesh versus ground-truth one.
Finally, end-to-end positioning accuracy for shortlisted body regions is measured in terms of horizontal $\ell_2$ distance between the predictions and the ground-truth scanner positions.

\subsection{Keypoint Positioning Accuracy}

\begin{table}[t]
    \centering
    \scalebox{0.7}{
    \begin{tabular}{c|p{0.85cm}|p{0.85cm}|p{0.85cm}|p{0.85cm}|p{0.85cm}|p{0.85cm}|p{0.85cm}|p{0.85cm}|p{0.85cm}|p{0.85cm}|p{0.85cm}|p{0.85cm}|p{0.85cm}}
    \toprule  
    Methods & R.Ak. & R.Kn. & R.H. & L.H. & L.Kn. & L.Ak. & R.Wr. & R.Eb. & R.Sh. & L.Sh. & L.Eb. & L.Wr. & Avg \\
    \midrule  
    PCK@0.1 & 94.6 & 91.8 & 87.0 & 88.5 & 90.7 & 94.2 & 85.5 & 89.5 & 90.7 & 90.1 & 86.7 & 77.9 & 88.9 \\
    \midrule  
    PCK@0.15 & 99.4 & 98.4 & 96.4 & 96.6 & 98.0 & 99.4 & 95.8 & 98.4 & 98.0 & 98.7 & 96.3 & 92.8 & 97.4\\
    \bottomrule  
    \end{tabular}}
    \caption{PCK@0.1 and PCK@0.15 evaluation of our proposed cross-modality 2D keypoint detector. We exclude hips, neck, and nose in the evaluation due to different  definitions of these joints in common 2D keypoint detectors \cite{coco2014} and SMPL predefined joint regressors \cite{loper2015smpl}}
    \label{table:2dpck}
    \vspace{-.5em}
\end{table}

\begin{table}[t]
    \centering
    \resizebox{\linewidth}{!}{
    \begin{tabular}{p{0.5cm}|p{0.85cm}|p{0.85cm}|p{0.85cm}|p{0.85cm}|p{0.85cm}|p{0.85cm}|p{0.85cm}|p{0.85cm}|p{0.85cm}|p{0.85cm}|p{0.85cm}|p{0.85cm}|p{0.85cm}}
    \toprule  
     Dir & R.Ak. & R.Kn. & L.Kn. & L.Ak. & R.Wr. & R.Eb. & R.Sh. & L.Sh. & L.Eb. & L.Wr. & H.T. & Avg \\
    \midrule
     x & 2.13 & 1.94 & 1.93 & 1.29 & 1.55 & 1.49 & 1.02 & 1.27 & 2.08 & 1.55 & 1.59 & 1.62\\
     \midrule
     y & 3.16 & 1.12 & 1.18 & 1.50 & 1.30 & 1.17 & 1.85 & 2.16 & 1.02 & 1.30 & 1.43 & 1.56 \\
     \midrule
     $l_2$ & 3.81 & 2.24 & 2.26 & 1.98 & 2.03 & 1.90 & 2.11 & 2.50 & 2.32 & 2.02 & 2.13 & 2.30 \\
    \bottomrule  
    \end{tabular}
    }
    \vspace{.5em}
    \caption{3D mesh regression reconstruction error (cm) on 108 data samples collected in lab settings.}
    \label{table:3d}
    \vspace{-1em}
\end{table}

From Table~\ref{table:2dpck}, we present the PCK@0.1 and PCK@0.15 of our proposed cross-modality-informed 2D body keypoint detector on our proprietary data (685 images) collected in lab settings, with heavy clutters and occlusions on the patients of interest to simulate real clinical environments. Table~\ref{table:2dpck} shows that our proposed dual-modality-informed keypoint detector is performing robustly under challenging scenarios with high accuracy, \eg, $>$90\% PCK@0.15 for all defined keypoints. Our 2D keypoint detector achieves an average accuracy of 9.61px in terms of MPJPE in image space, equivalent to 2.04cm in real-world coordinate system.

We further evaluate the proposed \textit{AutoCBCT} module by measuring the accuracy of the predicted 3D patient joints, as well as the accuracy of the mesh reconstruction. Results are shared in Table \ref{table:3d}, showing that our solution meets clinical accuracy requirement \wrt scanner positioning for these body key parts.

\begin{table}[t!]
    \centering
    \begin{tabular}
{c|c|c}
 \toprule
 Target location & Head Top & Radial Artery \\
  \midrule
 Mean Absolute Positioning Error (cm) & 0.86 & 2.34\\
 \bottomrule
\end{tabular}
    \caption{Positioning errors for Head Top and Radial Artery protocols. Evaluation is conducted in lab setting.}
    \label{table:endtoend}
    \vspace{-1em}
\end{table}

Finally we also include end-to-end positioning accuracy evaluation conducted in lab setting with respect to two common protocols ``Head Top" and ``Radial Artery", as shown in Table \ref{table:endtoend}. For this evaluation, the test subject (manikin) is posed differently on the patient support at different locations in the surgical room with various occlusion scenarios. In total 126 and 16 sets of data were studied for \textit{head top} and \textit{radial artery} protocols respectively. Positioning error are measured with laser measuring tools in X-Y plane.

\begin{figure}[t]
	\centering
	\includegraphics[width=.5\textwidth]{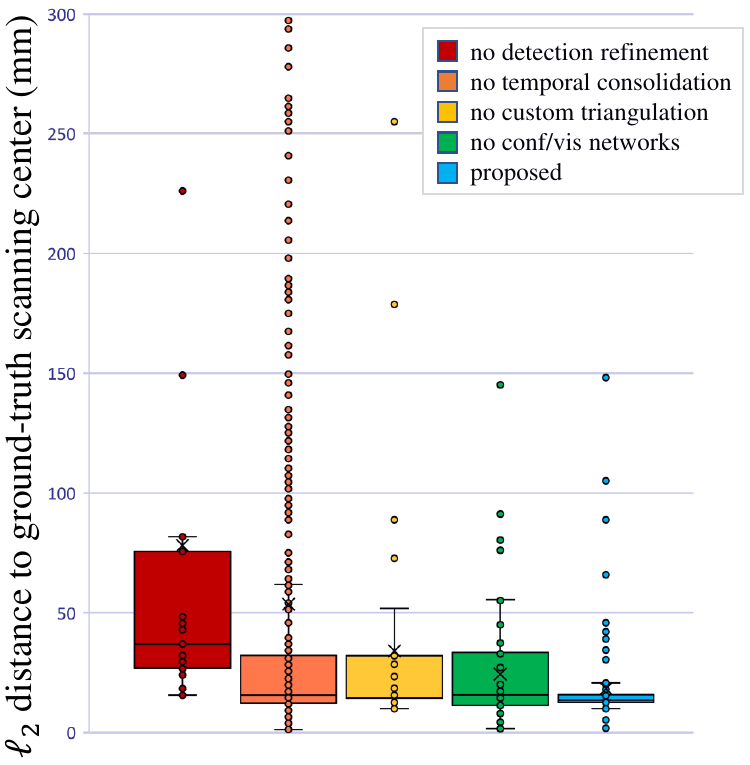}
	\caption{Ablation study in terms of accurate positioning of the right radial artery.}
	\label{fig:ablation_study}
\end{figure} 

\subsection{End-to-end Ablation Study}

To highlight the impact of our contributions to the positioning accuracy, we provide an ablation study in Fig. \ref{fig:ablation_study}, performed in terms of the correct in-plane localization of the right-wrist region for DSA purpose. 
We can especially observe the accuracy boost provided by our two-stage detection refinement, as well as the much higher consistency and robustness provided by our temporal aggregation module. Our proposed multi-stage weighted triangulation and  auxiliary prediction scoring further pushes our accuracy beyond the target clinical expectations.

\end{document}